\documentclass[10pt,twocolumn,letterpaper]{article}

\usepackage{cvpr}              

\usepackage[dvipsnames]{xcolor}

\usepackage{graphicx}
\usepackage{amsmath}
\usepackage{amssymb}
\usepackage{booktabs}
\usepackage{booktabs}
\usepackage{colortbl}
\usepackage{xcolor}         
\usepackage{CJKutf8}
\usepackage{makecell}

\definecolor{ForestGreen}{RGB}{34,139,34}
\newcommand{\fg}[1]{\textbf{\textcolor{ForestGreen}{#1}}}

\usepackage{pifont}
\newcommand{\cmark}{\ding{51}}%
\newcommand{\xmark}{\ding{55}}
\usepackage[accsupp]{axessibility} 

\usepackage{subcaption}

\definecolor{cvprblue}{rgb}{0.21,0.49,0.74}
\usepackage[pagebackref,breaklinks,colorlinks,citecolor=cvprblue]{hyperref}

\usepackage{amsmath,amsfonts,bm}

\def\eqref#1{equation~\ref{#1}}

\def\1{\bm{1}}

\DeclareMathAlphabet{\mathsfit}{\encodingdefault}{\sfdefault}{m}{sl}
\SetMathAlphabet{\mathsfit}{bold}{\encodingdefault}{\sfdefault}{bx}{n}

\newcommand{\Ls}{\mathcal{L}}

\definecolor{mycyan}{RGB}{212, 239, 251}
\usepackage{colortbl}
\usepackage{xcolor}
\definecolor{mygray}{gray}{.9}
\definecolor{goldenrod}{RGB}{245,245,220}
\newlength\savewidth\newcommand\shline{\noalign{\global\savewidth\arrayrulewidth\global\arrayrulewidth 1pt}\hline\noalign{\global\arrayrulewidth\savewidth}}
\newcolumntype{a}{>{\columncolor{mygray}}c}
\usepackage{fontawesome5}
\usepackage{booktabs}
\usepackage{makecell}
\usepackage{fontawesome5}
\usepackage{lipsum}
\usepackage{comment}
\usepackage{multirow}
\usepackage{wrapfig}
\usepackage{algorithm}
\usepackage{algorithmic}
\usepackage{xfrac}  

\usepackage{graphicx}
\usepackage{color}

\usepackage{amsthm}
\definecolor{darkgreen}{rgb}{0,0.7,0}

\definecolor{mygraytext}{gray}{.75}

\def\paperID{6204} 
\def\confName{CVPR}
\def\confYear{2024}

\title{FreeKD: Knowledge Distillation via Semantic Frequency Prompt}

\author{Yuan Zhang${}^{1}$, Tao Huang${}^{2}$, Jiaming Liu${}^{1}$, Tao Jiang${}^{3}$, Kuan Cheng${}^{1}$, Shanghang Zhang${}^{1\dag}$ \\
{\normalsize \textsuperscript{1} National Key Laboratory for Multimedia Information Processing,} \\ 
{\normalsize School of Computer Science,  Peking University}
{\normalsize \quad \textsuperscript{2}The University of Sydney \quad \textsuperscript{3}Zhejiang University }
}

\begin{document}
\maketitle

\renewcommand{\thefootnote}{\dag}
\footnotetext{Corresponding author. Shanghang Zhang is supported by the National Key Research and Development Project of China (No. 2022ZD0117801).}
\begin{abstract}
\vspace{-2mm}
Knowledge distillation (KD) has been applied to various tasks successfully, and mainstream methods typically boost the student model via spatial imitation losses. However, the consecutive downsamplings induced in the spatial domain of teacher model is a type of corruption, hindering the student from analyzing what specific information needs to be imitated, which results in accuracy degradation. To better understand the underlying pattern of corrupted feature maps, we shift our attention to the frequency domain. During frequency distillation, we encounter a new challenge: the low-frequency bands convey general but minimal context, while the high are more informative but also introduce noise. Not each pixel within the frequency bands contributes equally to the performance. To address the above problem: (1) We propose the Frequency Prompt plugged into the teacher model, absorbing the semantic frequency context during finetuning. (2) During the distillation period, a pixel-wise frequency mask is generated via Frequency Prompt, to localize those pixel of interests (PoIs) in various frequency bands. Additionally, we employ a position-aware relational frequency loss for dense prediction tasks, delivering a high-order spatial enhancement to the student model. We dub our \textbf{Fre}qu\textbf{e}ncy \textbf{K}nowledge \textbf{D}istillation method as \textbf{FreeKD}, which determines the optimal localization and extent for the frequency distillation. Extensive experiments demonstrate that FreeKD not only outperforms spatial-based distillation methods consistently on dense prediction tasks (e.g., FreeKD brings \textbf{3.8} AP gains for RepPoints-R50 on COCO2017 and \textbf{4.55} mIoU gains for PSPNet-R18 on Cityscapes), but also conveys more robustness to the student. Notably, we also validate the generalization of our approach on large-scale vision models (e.g., DINO and SAM). Code is available at \href{https://github.com/Gumpest/FreeKD}{here}.
\end{abstract}    
\vspace{-3mm}
\section{Introduction}
\label{sec:intro}

\begin{figure}[t]
    \centering
    \includegraphics[width=0.45\textwidth]{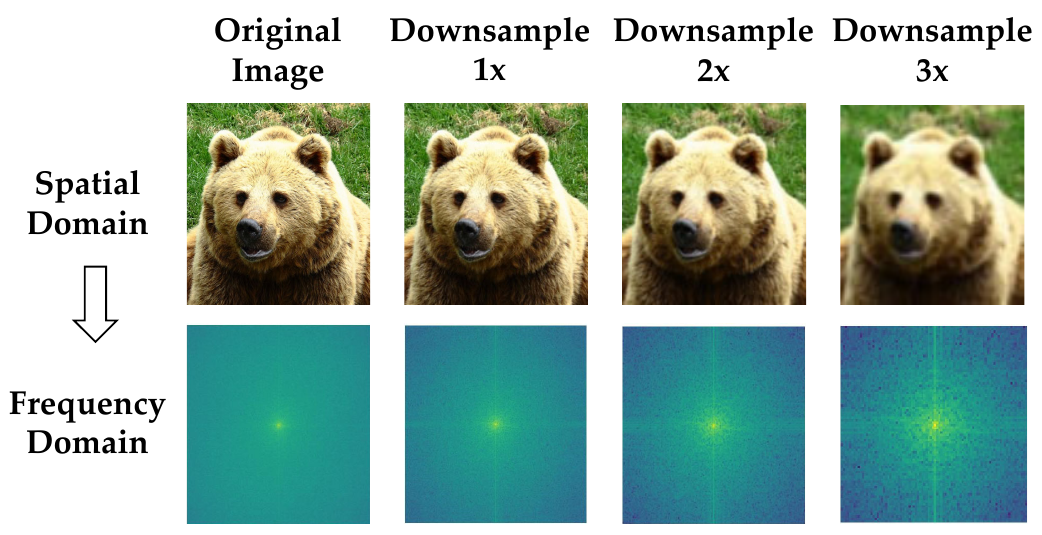}
    \caption{\textbf{Comparison of the presentation of the bear at different downsampling ratios on spatial and frequency domain.}}
    \label{fig:Moti}
    \vspace{-3mm}
\end{figure}

In the quest for significant advancements, recent deep learning models have witnessed a substantial increase in both depth and width, as exemplified by notable works such as \cite{he2016deep, liu2021swin, kirillov2023segment}. However, pursuing larger and more powerful models results in unwieldy and inefficient deployments on resource-limited edge devices. To address this dilemma, knowledge distillation (KD) \cite{hinton2015distilling, Du2021DistillingOD, dai2021general, yang2022focal, huang2023knowledge} has emerged as a promising solution to transfer the knowledge encapsulated within a heavy model (teacher) to a more compact, pocket-size model (student).

Among diverse computer vision tasks, the transfer of dark knowledge for dense prediction tasks poses unique challenges, particularly requiring fine-grained distillation at the feature level. Recent distillation methods have aimed to enhance performance through spatial-level distillation losses, refining valuable information within the features. However, the sequential downsampling applied in the spatial domain of the teacher model introduces a form of corruption. This corruption hampers the student's ability to discern specific information that should be mimicked, resulting in a decline in accuracy.

\begin{figure}[t]
    \centering
    \includegraphics[width=0.42\textwidth]{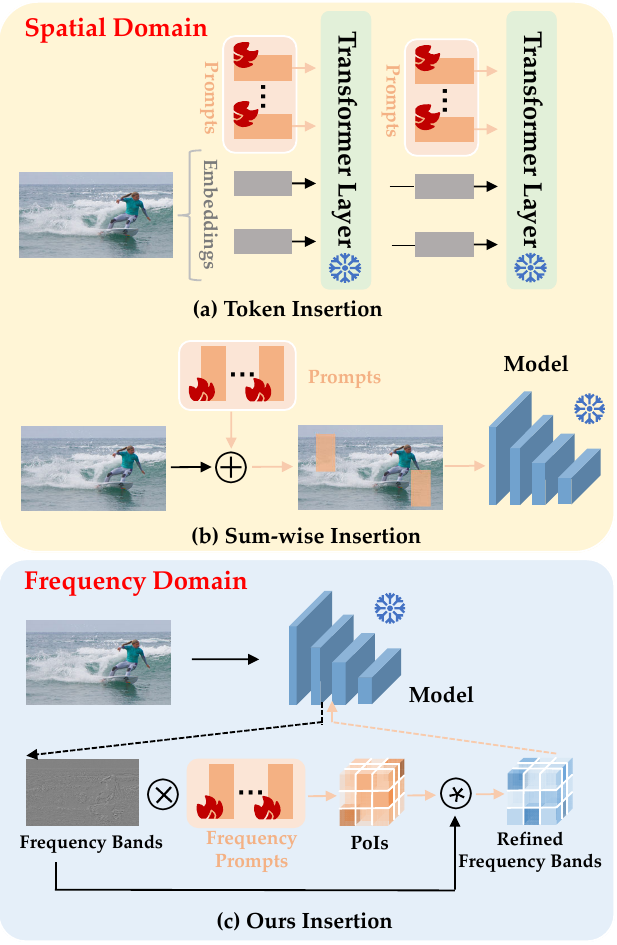}
    \caption{\textbf{Comparisons with other insertion methods of spatial prompts.} (a) Prompts are inserted into the encoder layer as tokens. (b) Sum-wise on RGB channels of input image. (c) Ours interact with intermediate features. Best view in color.}
    \label{fig:Prompts}
    \vspace{-2mm}
\end{figure}

As illustrated in Figure \ref{fig:Moti}, downsampling operations prominently remove high-frequency image details in the frequency domain, revealing underlying patterns not easily discernible from raw pixel values \cite{xu2020learning, chen2020frequency, xie2022masked}. This observation prompts us to explore the potential of leveraging frequency signals for knowledge distillation. However, directly employing this approach raises two significant challenges: \textbf{(a)} The low-frequency bands from the teacher model convey general yet minimal contextual information, characterized by smooth variations \cite{xiang2008invariant, zhu2017application}. If the student is forced to imitate all pixels of low-frequency bands directly, it tends to focus on easy but less informative samples, aiming to reduce loss. \textbf{(b)} The high-frequency range provides more fine-grained and distinctive signals, with salient transitions enhancing the student's robustness and generalizability \cite{zhang2022wavelet}. However, when the student mimics high-frequency pixels, it also captures noise, leading to undesired degradation. Therefore, the challenge lies in localizing worthy pixels of interest (PoIs) in both frequency bands.
To address these challenges, we introduce the semantic Frequency Prompt as depicted in Figure \ref{fig:Prompts} (c). Initially, a set of Points of Interest (PoIs) masks is generated by encoding similarities between prompts and frequency bands. Subsequently, the masked frequency bands, rather than the vanilla ones, are supervised by task loss. This approach provides precise guidance for the student in reconstructing the teacher's frequency bands — a crucial aspect of knowledge distillation. Importantly, the Frequency Prompt differs from previous spatial prompts in both insertion method and the transferred substance. In Figure \ref{fig:Prompts}, Prompt Tokens (VPTs) \cite{jia2022visual, zhu2023visual} are inserted as tokens for transformer series tasks, while Contrastive Texture Attention Prompts (CTAP) \cite{gan2023decorate} are summed point by point on the input image, avoiding occlusion. In contrast, the localization of our Frequency Prompts is flexible, depending on where the student intends to imitate.
This involves incorporating a position-aware relational frequency loss, where positional channel-wise weights are derived from cross-layer information. These weights act as an adaptive gating operation, selectively choosing relevant channels from frequency bands.

With the above key designs, we propose a \textbf{Fre}qu\textbf{e}ncy \textbf{K}nowledge \textbf{D}istillation pipeline called \textbf{FreeKD}, where the student is under fine-grained frequency imitation principle. Extensive experimental results show that our method surpasses current state-of-the-art spatial-based methods consistently in standard settings of object detection and semantic segmentation tasks. For instance, FreeKD obtains $42.4$ AP with RepPoints-R50 student on the COCO dataset, surpassing DiffKD \cite{huang2023knowledge} by $0.7$ AP; while on semantic segmentation, FreeKD outperforms MGD \cite{yang2022masked} by $0.8\%$ with PSPNet-R18 student on Cityscapes test set. Moreover, we implement FreeKD on large-scale vision model settings, and our method significantly outperforms the baseline method. Finally, we are surprised that the student distilled by FreeKD exhibits better domain generalization capabilities (e.g.,  FreeKD outperforms DiffKD by $1.0\%$ rPC \cite{wang2022generalizing}). 

In a nutshell, the contributions of this paper are threefold:
\begin{enumerate} 
\item We introduce a novel knowledge distillation manner (FreeKD) from the frequency domain, and make the first attempt to explore its potential for distillation on dense prediction tasks, which breaks the bottleneck of spatial-based methods.
\item To the best of our knowledge, we are the first to propose Frequency Prompt especially for frequency knowledge distillation, eliminating unfavorable information from frequency bands, and a position-aware relational frequency loss for dense prediction enhancement.
\item We validate the effectiveness of our method through extensive experiments on various benchmarks, including large-scale vision model settings. Our approach consistently outperforms existing spatial-based methods, demonstrating significant improvements and enhanced robustness in students distilled by FreeKD.
\end{enumerate} 

\section{Related Work}

\subsection{KD on Dense Prediction Tasks}
In recent years, knowledge distillation for dense prediction tasks such as object detection and semantic segmentation has garnered significant attention, owing to its practical applications and the inherent challenges of distilling fine-grained pixel-level recognition and localization features. Early approaches~\cite{chen2017learning,li2017mimicking} primarily concentrated on distilling classification and regression outputs or intermediate features using traditional loss functions such as Kullback–Leibler divergence and mean square error. However, recent research has shifted its focus towards mimicking valuable information while filtering out noisy features in the intermediate dense representations. This shift is driven by the observation that dense features often contain redundant information, which can burden the student model. To address this, contemporary works employ techniques like generating pixel-level masks based on ground-truth boxes~\cite{wang2019distilling,guo2021distilling}, leveraging feature attentions~\cite{shu2021channel,yang2021focal}, and introducing learnable mask tokens~\cite{huang2022masked} for feature refinement. Besides, some approaches propose to reducing the representation gap between teacher and student via normalizing the features with Pearson correlation~\cite{cao2022pkd} or denoising the features with diffusion models~\cite{huang2023knowledge}.  However, the consecutive downsamplings induced in the spatial domain of the teacher model is a type of corruption, hindering the student from analyzing what specific information needs to be imitated, which results in accuracy degradation. To better understand the underlying pattern of corrupted feature maps, we shift our attention to the frequency domain.

\subsection{Frequency Analysis Methods}
Frequency domain analysis has found extensive application in various computer vision tasks, including image classification~\cite{xu2020learning,williams2018wavelet}, image generation~\cite{jiang2021focal}, and image super-resolution~\cite{pang2020fan}. Early studies \cite{oppenheim1981importance,piotrowski1982demonstration,hansen2007structural} indicate that in the frequency domain, the phase component predominantly captures high-level semantics of the original signals, while the amplitude component retains low-level statistics. Consequently, underlying image patterns are more conveniently observed in the frequency representation compared to raw pixel values in the spatial domain. In this context, wavelet analysis stands out as a particularly effective method in image processing~\cite{mallat1999wavelet,fujieda2017wavelet,zhang2022wavelet}, as it can capture multiscale frequency domain information in a compact representation. Unlike other frequency analysis methods like Fourier analysis, wavelet analysis offers a more comprehensive perspective. Leveraging wavelet analysis, our method is tailored for dense prediction tasks, demonstrating superior distillation on image patterns when compared to distilling raw pixel values in the spatial domain.

\section{Proposed Approach: FreeKD}
\label{sec:method}

\begin{figure*}[t]
    \centering
    \includegraphics[width=1\textwidth]{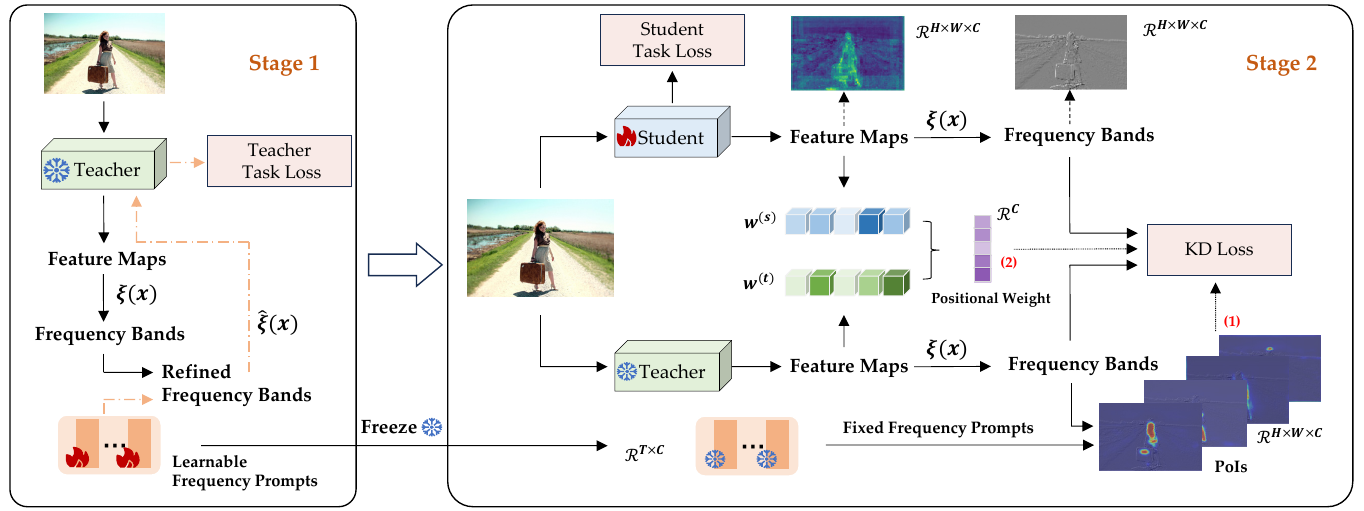}
    \caption{\textbf{Overview of our FreeKD pipeline.} The pipeline includes two stages. \textbf{Stage 1}: Frequency prompts make interaction with intermediate frequency bands, and are supervised by the teacher task loss. \textbf{Stage 2}: First, the distillation feature maps of student and teacher transform into the frequency domain, respectively. Then, receiving frequency prompts from stage 1, we request the frozen ones multiply with teacher frequency bands, and generate the PoIs of bands. Finally, a channel-wise positional-aware weight is determined by the teacher spatial gate and student gate together. The flow (1) in the figure decides where to distill and flow (2) indicates the extent of the distillation.}
    \label{fig:Arch}
    \vspace{-2mm}
\end{figure*}

In this section, we first demonstrate vanilla knowledge distillation via frequency loss. To further provide more precise PoIs, we design a novel Frequency Prompt to generate pixel imitation principles. Finally, we propose a position-aware relational loss to enhance the sensitivity to dense prediction. The architecture of FreeKD is illustrated in Figure \ref{fig:Arch}.

\subsection{Distillation with Frequency}
Dilations and translations of the Mother function $\Phi(t)$, define an orthogonal wavelet basis:
\begin{equation} \label{eq:motherfunction}
    \Phi_{(s, d)}(t) = 2^{\frac{s}{2}} \Phi(2^{s} t - d), \quad s, d \in \bm{Z}
\end{equation}
where $\bm{Z}$ is the set of all integers and the factor $\frac{s}{2}$ maintains a constant norm independent of scale $s$. The variables $s$ and $d$, scales and dilates the mother function $\Phi$ to generate wavelets in $\mathcal{L}_2$ spaces. To create frequency representations, Discrete Wavelet Transformation (DWT) $\xi$ is applied for frequency bands decomposition via $\Phi$ to each channel as follows:
\begin{equation} \label{eq:dwt}
     \mathcal{B}_l = \xi(x),
\end{equation}
where $l$ is the decomposition level. When the level is set to $1$, the feature map $\bm{F} \in\mathbb{R}^{C\times H\times W}$ can be split into four bands, and $\mathcal{B}_1 = \{\texttt{LL}, \texttt{HL}, \texttt{LH}, \texttt{HH}\}$, where $\texttt{LL}$ indicates the low-frequency band ($\bm{R_{\texttt{LL}}}\in\mathbb{R}^{C\times H_{\texttt{LL}}\times W_{\texttt{LL}}}$ represents its corresponding tensor), and the others are high-frequency bands. When $l$ is $2$, the \texttt{LL} band can be further decomposed into \texttt{LL2}, \texttt{HL2}, \texttt{LH2} and \texttt{HH2}. In this paper, we set $l = 3$ for all distillation experiments.

In order to learn dark knowledge of the teacher, one typical manner is to mimic the tensor pixel-wisely. Regularly, $\bm{F}^{(t)}\in\mathbb{R}^{C\times H\times W}$ and $\bm{F}^{(s)}\in\mathbb{R}^{C_s\times H\times W}$ denote the feature maps of teacher and student networks respectively, and the frequency bands imitation can be fulfilled via: 
\begin{equation} \label{eq:mimic}
    \Ls_\mathrm{FKD} = \sum_{k=1}^L \left\| a_{k} - b_{k} \right\|_1 ,
    \nonumber
\end{equation}
\begin{equation} 
    a_{k} \in \xi(\bm{F}^{(t)}), b_{k} \in \xi(\phi(\bm{F}^{(s)})),
\end{equation}
where $L$ is the number of frequency bands, and $\phi$ is a linear projection layer to adapt $\bm{F}^{(s)}$ to the same resolution as $\bm{F}^{(t)}$. The student model studies general laws via low-frequency imitation, and salient pattern (including fine textures, edges, and \textit{noise}) from the high-frequency.

\subsection{Semantic Frequency Prompt}
Therefore, we introduce a learnable frequency prompt $\mathcal{P}\in\mathbb{R}^{B\times T\times C}$ to deliver $T$ pixel imitation principles in $C$ channels of $B$ frequency bands, and it will finetune the teacher model first. For simplicity, we choose the frequency band \texttt{HH} from $B$ bands and its corresponding prompt $\bm{P}\in\mathbb{R}^{ T\times C}$ as an example, and the rest are the same. 

Unlike previous insertion methods of spatial-based prompts, our approach requests the frequency prompt to interact with the band, a better way to know the manifolds embedded in frequency spaces. In this paper, we adopt the matrix multiplication manner to calculate the mutual information $\bm{M} \in\mathbb{R}^{C\times H_{\texttt{HH}} W_{\texttt{HH}}}$ between prompt $\bm{P}$ and frequency pixels $\bm{R}^{(t)}$ in the teacher band:
\begin{equation}
    \bm{M} = \bm{P} \times \bm{R}^{(t)},
\end{equation}
where we flatten the band $\texttt{HH}$ into shape $(C, H_{\texttt{HH}}\times W_{\texttt{HH}})$ to fit matrix multiplication. 

Then, to connect with the task loss $\Ls_\mathrm{finetune}$ supervision and support stochastic gradient descent, a masked frequency band is utilized to substitute the original band $\texttt{HH}$:
\begin{equation}
    \hat{\bm{R}^{(t)}} = \sum_{i=1}^T \sigma {(\bm{M}_i)} \circledast \bm{R}^{(t)} ,
\end{equation}
where we turn the mutual information $\bm{M}$ into a probability space to function as the masks. The symbol $\sigma$ denotes the sigmoid function and $\circledast$ means element-wise multiplication. 

After collecting all $B$ masked frequency bands, we perform an Inverse Discrete Wavelet Transformation (IDWT) $\tilde{\xi}$ on them to the spatial domain:
\begin{equation}
   \hat{ \bm{F}^{(t)}} = \tilde{\xi}(\hat{\mathcal{B}_l}),
\end{equation}
and we send the new feature map $\hat{ \bm{F}^{(t)}}$ back to the teacher model. The finetune loss can be treated as an observation of mask quality, and minimize to force the frequency prompts to focus on the substantial pixels of the band.

However, simply minimizing $\Ls_\mathrm{finetune}$ would lead to an undesired collapse of the $T$ sets of masks generated by the frequency prompt. Specifically, some masks will be learned to directly recover all the bands, filled with $1$ everywhere. To make the prompt represent $T$ sets PoIs of the band, we propose a Prompt-Dissimilarity loss based on the Jaccard coefficient:
\begin{equation}
    \Ls_\mathrm{dis} = \frac{1}{T^2}\sum_{i=1}^{T}\sum_{j=1}^{T} \Theta_\mathrm{Jaccard}(\bm{M}_i, \bm{M}_j)
\end{equation}
with
\begin{equation}
    \Theta_\mathrm{Jaccard} (\bm{m}, \bm{n}) = \frac{{|\bm{m} \cap \bm{n}|}}{{|\bm{m} \cup \bm{n}|}} ,
\end{equation}
where $\bm{m}\in\mathbb{R}^{N}$ and $\bm{n}\in\mathbb{R}^{N}$ are two vectors. Jaccard loss is widely used to measure the degree of overlap between two masks in segmentation tasks. By minimizing the coefficients of each mask pair, we can make masks associated with different PoIs. As a result, the training loss of prompt is composed of finetune loss and dissimilarity loss:
\begin{equation}
    \Ls_\mathrm{prompt} = \Ls_\mathrm{finetune} + \lambda\Ls_\mathrm{dis} ,
\end{equation}
where $\lambda$ is a factor for balancing the loss. In this paper, we set $\lambda=1$ for all distillation experiments and allocate $T = 2$ imitation principles for each frequency band, as the frequency prompt is easier to converge (e.g., the teacher \textit{FCOS ResNet101} has $40.8$ mAP on COCO val set, and the finetuned one is $39.9$). Notably, we still utilize the original teacher instead of the finetuned one to distill for the students for the fairness.

\subsection{Position-aware Relational Loss}
With the help of frequency prompt, we can already localize the PoIs of bands to improve the performance of frequency distillation. As frequency responses come from a local region, encoding original features with positional importance is thus necessary to distinguish the objects for dense prediction. Hence we introduce the Position-aware Relational Loss to provide high-order spatial enhancement for the student model. First,  the relational attention from multi-receptive fields can be represented as:
\begin{equation}
    \bm{A} = Softmax(\psi (\bm{F}) \bm{F}^{T}) ,
\end{equation}
where $\psi (\bm{F})$ denotes the spatial feature of the latter layer than $\bm{F}$. Thus $\bm{A} \in \mathbb{R}^{C \times C}$ serves as a bridge to find the position-aware correlations across different layers. Then, the gating operation is generated based on the spatial perceptions to form the position-aware loss relation weight:
\begin{equation}
    \omega = \mathcal{G}(\bm{A}) \in \mathbb{R}^{1 \times C},
\end{equation}
where $\mathcal{G}$ denotes the gating weight generated by a Multilayer Perceptron (MLP). Therefore, Eq. \ref{eq:mimic} can be reformulated as:
\begin{equation} \label{eq:re_loss}
    \Ls_\mathrm{FKD} = \sum_{k=1}^L \omega^{(r)} \left\| a_{k} - b_{k} \right\|_1 ,
\end{equation}
with $\omega^{(r)} = \omega^{(t)} \circledast \omega^{(s)}$ generated by the teacher and student position-aware relation weight. The reason is that the channels in distillation should consist of the ones both meaningful to the teacher and student. Our eventual
frequency distillation loss can be formulated as:
\begin{small}
\begin{equation} \label{eq:freekd}
    \Ls_\mathrm{FreeKD} = \sum_{k=1}^L \omega^{(r)} \left\| \bm{M} {\circledast} a_{k} - \bm{M} \circledast b_{k} \right\|_1.
\end{equation}
\end{small}

\subsection{Overall loss}
To sum up, we train the student detector with the total loss formulated as:
\begin{equation} \label{eq:overall_loss}
    \Ls_\mathrm{student} = \Ls_\mathrm{task} + \mu \Ls_\mathrm{FreeKD} ,
\end{equation}
where $\mu$ is a factor for balancing the losses. The distillation loss is applied to intermediate feature maps (e.g., the feature pyramid network \cite{lin2017feature} (FPN) in object detection tasks), so it can be easily applied to different architectures.
\section{Experiments}
In this paper, to validate the superiority of our method, we conduct extensive experiments on object detection and semantic segmentation tasks, with various model architectures (including CNN-based and Transformer-based). Furthermore, we evaluate the robustness of detectors trained with FreeKD on the COCO-C benchmark, to exhibit its better domain generalization capabilities.

\subsection{Object Detection}

\definecolor{mygray}{gray}{.9}
\begin{table*}[t]
    \centering
    \begin{minipage}[t]{0.48\textwidth}
	\renewcommand\arraystretch{1.3}
	\setlength\tabcolsep{1.3mm}
	\centering
	\caption{\textbf{Object detection performance via FreeKD in baseline settings on COCO val set.}}
	\vspace{-1mm}
	\label{tab:det}
    \footnotesize
	\begin{tabular}{p{1.6cm}p{1.4cm}|p{1.4cm}cccc}
	    \shline
	    \multicolumn{2}{c|}{Method} & \makecell[c]{AP} & AP$_{S}$ & AP$_{M}$ & AP$_{L}$\\
	    \shline
	    \multicolumn{6}{c}{\textit{One-stage detectors}}\\
	    \multicolumn{2}{l|}{T: RetinaNet-R101} & 38.9 & 21.0 & 42.8 & 52.4\\
	    \multicolumn{2}{l|}{S: RetinaNet-R50} & 37.4 & 20.0 & 40.7 & 49.7 \\
        FRS \cite{2021Distilling} & \texttt{\scriptsize{NeurlPS21}} &39.3 (1.9$\uparrow$)& 21.5 & 43.3 & 52.6\\
        FGD \cite{yang2021focal} & \texttt{\scriptsize{CVPR22}} & 39.6 (2.2$\uparrow$) & \textbf{22.9} & 43.7 & 53.6 \\
	    DiffKD \cite{huang2023knowledge} & \texttt{\scriptsize{NeurlPS23}} & 39.7 (2.3$\uparrow$) & 21.6 & 43.8 & 53.3\\
        \rowcolor{mygray}
        FreeKD & &\fg{{39.9}} (\textbf{\fg{2.5$\uparrow$}}) & 21.2 & \textbf{44.0} & \textbf{53.7}\\
	    \hline
	    \multicolumn{6}{c}{\textit{Two-stage detectors}}\\
	    \multicolumn{2}{l|}{\scriptsize{T: Faster RCNN-R101}} & 39.8 & 22.5 & 43.6 & 52.8 \\
	    \multicolumn{2}{l|}{\scriptsize{S: Faster RCNN-R50}} & 38.4 & 21.5 & 42.1 & 50.3\\
        FRS \cite{2021Distilling} & \texttt{\scriptsize{NeurlPS21}} &39.5 (1.1$\uparrow$) & 22.3 & 43.6 & 51.7\\
        FGD \cite{yang2021focal} & \texttt{\scriptsize{CVPR22}} & 40.4 (2.0$\uparrow$) & 22.8 & 44.5 & 53.5 \\
	    DiffKD \cite{huang2023knowledge} & \texttt{\scriptsize{NeurlPS23}} & 40.6 (2.2$\uparrow$) & 23.0 & 44.5 & 54.0 \\
        \rowcolor{mygray}
        FreeKD & &\fg{{40.8}} (\textbf{\fg{2.4$\uparrow$}}) & \textbf{23.1} & \textbf{44.7} & \textbf{54.0}\\
	    \hline
	    \multicolumn{6}{c}{\textit{Anchor-free detectors}}\\
        \multicolumn{2}{l|}{T: FCOS-R101} & 40.8 & 24.2 & 44.3 & 52.4\\
        \multicolumn{2}{l|}{S: FCOS-R50} & 38.5 & 21.9 & 42.8 & 48.6\\
        FRS \cite{2021Distilling} & \texttt{\scriptsize{NeurlPS21}} &40.9 (2.4$\uparrow$) & 25.7 & 45.2 & 51.2\\
        FGD \cite{yang2021focal} & \texttt{\scriptsize{CVPR22}} & 42.1 (3.6$\uparrow$) & \textbf{27.0} & 46.0 & 54.6\\
        DiffKD \cite{huang2023knowledge} & \texttt{\scriptsize{NeurlPS23}} & 42.4 (3.9$\uparrow$) & 26.6 & 45.9 & 54.8\\
        \rowcolor{mygray}
        FreeKD & &\fg{{42.9}} (\textbf{\fg{4.4$\uparrow$}}) & 26.8 & \textbf{46.8} & \textbf{55.4}\\
	    \shline
	\end{tabular}
    \end{minipage}
    \hspace{2mm}
    \begin{minipage}[t]{0.48\textwidth}
	\renewcommand\arraystretch{1.3}
	\setlength\tabcolsep{1.3mm}
	\centering
    \caption{\textbf{Object detection performance via FreeKD in stronger settings on COCO val set.} \textit{CM RCNN}: Cascade Mask RCNN.}
	\vspace{-1mm}
	\label{tab:det2}
    \footnotesize
	\begin{tabular}{p{1.6cm}p{1.4cm}|cccccc}
	    \shline
	    \multicolumn{2}{c|}{Method} & \makecell[c]{AP} & AP$_{S}$ & AP$_{M}$ & AP$_{L}$\\
	    \shline
	    \multicolumn{6}{c}{\textit{One-stage detectors}}\\
	    \multicolumn{2}{l|}{T: RetinaNet-X101} & 41.2  & 24.0 & 45.5 & 53.5\\
	    \multicolumn{2}{l|}{S: RetinaNet-R50} & 37.4 & 20.0 & 40.7 & 49.7 \\
        FRS \cite{2021Distilling} & \texttt{\scriptsize{NeurlPS21}} &40.1 (2.7$\uparrow$) & 21.9 & 43.7 & 54.3\\
        FGD \cite{yang2021focal} & \texttt{\scriptsize{CVPR22}} & 40.7 (3.3$\uparrow$) & \textbf{22.9} & 45.0 & 54.7\\
	    DiffKD \cite{huang2023knowledge} & \texttt{\scriptsize{NeurlPS23}} & 40.7 (3.3$\uparrow$) & 22.2 & 45.0 & 55.2\\
        \rowcolor{mygray}
        FreeKD & &\fg{{41.0}} (\textbf{\fg{3.6$\uparrow$}}) & 22.3 & \textbf{45.1} & \textbf{55.7}\\
	    \hline
	    \multicolumn{6}{c}{\textit{Two-stage detectors}}\\
	    \multicolumn{2}{l|}{\scriptsize{T: CM RCNN-X101}} & 45.6 & 26.2 & 49.6 & 60.0 \\
	    \multicolumn{2}{l|}{\scriptsize{S: Faster RCNN-R50}} & 38.4 & 21.5 & 42.1 & 50.3 \\
        CWD~\cite{shu2021channel} & \texttt{\scriptsize{ICCV21}} &41.7 (3.3$\uparrow$) & 23.3 & 45.5 & 55.5\\
        FGD \cite{yang2021focal} & \texttt{\scriptsize{CVPR22}} & 42.0 (3.6$\uparrow$) & 23.7 & 46.4 & 55.5\\
	    DiffKD \cite{huang2023knowledge} & \texttt{\scriptsize{NeurlPS23}} & 42.2 (3.8$\uparrow$) & \textbf{24.2} & 46.6 & 55.3 \\
        \rowcolor{mygray}
        FreeKD & & \fg{42.4} (\textbf{\fg{4.0$\uparrow$}}) & 24.1 & \textbf{46.7} & \textbf{55.9}\\
	    \hline
	    \multicolumn{6}{c}{\textit{Anchor-free detectors}}\\
        \multicolumn{2}{l|}{T: RepPoints-X101} & 44.2 & 26.2 & 48.4 & 58.5\\
        \multicolumn{2}{l|}{S: RepPoints-R50} & 38.6 & 22.5 & 42.2 & 50.4\\
        FKD \cite{zhang2020improve} & \texttt{\scriptsize{ICLR20}} & 40.6 (2.0$\uparrow$) & 23.4 & 44.6 & 53.0\\
        FGD \cite{yang2021focal} & \texttt{\scriptsize{CVPR22}} & 41.3 (2.7$\uparrow$) & \textbf{24.5} & 45.2 & 54.0\\
        DiffKD \cite{huang2023knowledge} & \texttt{\scriptsize{NeurlPS23}} & 41.7 (3.1$\uparrow$) & 23.6 & 45.4 & 55.9\\
        \rowcolor{mygray}
        FreeKD & & \fg{{42.4}} (\textbf{\fg{3.8$\uparrow$}}) & 24.3 & \textbf{46.4} & \textbf{56.6}\\
	    \shline
	\end{tabular}
    \end{minipage}
    \vspace{-2mm}
\end{table*}
\subsubsection{\textbf{Datasets.}}
We experiment on MS COCO detection dataset \cite{lin2014microsoft}, which contains 80 object classes. We train the student models on COCO \texttt{train2017} set and evaluate them with average precision (AP) on \texttt{val2017} set.

\subsubsection{\textbf{Network Architectures.}}
Our evaluation includes two-stage models \cite{ren2015faster}, anchor-based one-stage models \cite{lin2017focal}, as well as anchor-free one-stage models \cite{tian2019fcos,yang2019reppoints}, to validate the efficacy of FreeKD across diverse detection architectures.

\label{sec:od-im}

\subsubsection{\textbf{Implementation Details.}}
For the object detection task, we conduct feature distillation on the predicted feature maps sourced from teacher's neck. We adopt ImageNet pretrained backbones and inheriting strategy following previous KD works \cite{yang2022masked, huang2023knowledge, zhang2023avatar} during training. All the models are trained with the official strategies (SGD, weight decay of 1e-4) of 2X schedule in MMDetection \cite{chen2019mmdetection}. We train the student with our FreeKD loss $\Ls_\mathrm{FreeKD}$, regression KD loss, and task loss for the object detection task. Concretely, the loss weights $\mu$ of $\Ls_\mathrm{FreeKD}$ in Eq.\ref{eq:overall_loss} on \textit{Faster RCNN}, \textit{RetinaNet}, \textit{FCOS}, and \textit{RepPoints} are $1$, $5$, $10$, and $10$.

\subsubsection{\textbf{Experimental Results.}}
\textbf{Results on baseline settings.}
Our results compared with previous methods are summarized in Table \ref{tab:det}, where we take ResNet-101 (R101) \cite{he2016deep} backbone as the teacher network, and ResNet-50 (R50) as the student. Our FreeKD can significantly improve the performance of student models over their teachers on various network architectures. For instance, FreeKD improves FCOS-R50 by $4.4$ AP and surpasses DiffKD \cite{huang2023knowledge} by $0.5$ AP. Besides, FreeKD benefits more to detecting large-size objects (AP$_{L}$), as larger objects would involve more frequency bands and cross-domain information.

\textbf{Results on stronger settings.}
We further investigate our efficacy on stronger teachers whose backbones are replaced by stronger ResNeXt (X101) \cite{xie2017aggregated}. The results in Table \ref{tab:det2} demonstrate that student detectors achieve more enhancements with our FreeKD, especially when with a RepPoints-X101 teacher, FreeKD gains a substantial improvement of $3.8$ AP over the RepPoints-R50. Additionally, our method outperforms existing KD methods by a large margin, and the improvement of FreeKD compared to DiffKD \cite{huang2023knowledge} is greater for all cases than the improvement of DiffKD \cite{huang2023knowledge} compared to FGD \cite{yang2022focal}.

\subsection{Semantic segmentation}
\definecolor{mygray}{gray}{.9}
\begin{table}[t]
    \centering
    \setlength{\tabcolsep}{2.5pt}
    \renewcommand{\arraystretch}{1.5}
    \footnotesize
	\centering
	\caption{\textbf{Semantic segmentation performance via FreeKD on Cityscapes val set.} FLOPs is measured based on an input image size of 512 × 512.}
     \vspace{-1mm}
	\label{tab:seg}
    \begin{tabular}{p{1.5cm}p{0.9cm}|c|c|c}
        \shline
        \multicolumn{2}{c|}{Method} & Params (M) & FLOPs (G) & {mIoU (\%)} \\
        \shline
        \multicolumn{2}{l|}{T: PSPNet-R101} &  70.43 & 574.9 & 78.34\\
        \hline
        \multicolumn{2}{l|}{\scriptsize{S: PSPNet-R18}}  & \multirow{4}*{13.1} & \multirow{4}*{125.8} & 69.85 \\
        CWD \cite{shu2021channel} & \texttt{\scriptsize{ICCV21}} & ~ & ~ & 73.53 \\
        MGD \cite{yang2022masked} & \texttt{\scriptsize{ECCV22}} & ~ & ~ & 73.63 \\
        \rowcolor{mygray}
	    FreeKD & &  ~ & ~ & \textbf{74.40}   \\
        \hline
        \multicolumn{2}{l|}{\scriptsize{S: DeepLabV3-R18}}  & \multirow{4}*{12.6} & \multirow{4}*{123.9} & 73.20 \\
        CWD \cite{shu2021channel} & \texttt{\scriptsize{ICCV21}} & ~ & ~ & 75.93 \\
        MGD \cite{yang2022masked} & \texttt{\scriptsize{ECCV22}} & ~ & ~ & 76.02 \\
        \rowcolor{mygray}
	    FreeKD & &  ~ & ~ & \textbf{76.45} \\
        \shline
	\end{tabular}
     \vspace{-2mm}
\end{table}
\subsubsection{\textbf{Datasets.}}
 We conduct experiments on Cityscapes dataset \cite{Cordts2016Cityscapes} to valid the effects of our method, which contains 5000 high-quality images (2975, 500, and 1525 images for the training, validation, and testing). We evaluate all the student networks with mean Intersection-over-Union (mIoU).
 
\subsubsection{\textbf{Network architectures.}}
For all segmentation experiments, we take PSPNet-R101 \cite{zhao2017pyramid} as the teacher network. While for the students, we use various frameworks (DeepLabV3 \cite{chen2018encoder} and PSPNet) with ResNet-18 (R18) to demonstrate the efficacy of our method.

\label{sec:ss-im}

\subsubsection{\textbf{Implementation Details.}}
For the semantic segmentation task, we conduct feature distillation on the predicted segmentation maps. All the models are trained with the official strategies of 40K iterations schedule with 512 × 512 input size in MMSegmentation \cite{mmseg2020}, where the optimizer is SGD and the weight decay is 5e-4.
A polynomial annealing learning rate scheduler is adopted with an initial value of 0.02.
\subsubsection{\textbf{Experimental results.}}
The experimental results are summarized in \ref{tab:seg}. FreeKD further improves the performance of state-of-the-art MGD \cite{yang2022masked} on both homogeneous and heterogeneous settings. For instance, the ResNet-18-based PSPNet gets $0.77$ mIoU gain and that based DeepLabV3 gets $0.43$ mIoU.

\subsection{Natural Corrupted Augmentation}
\definecolor{mygray}{gray}{.9}
\begin{table}[t]
    \centering
    \setlength{\tabcolsep}{10pt}
    \renewcommand{\arraystretch}{1.5}
    \footnotesize
	\centering
	\caption{\textbf{Performance of robust object detection via FreeKD on COCO-C dataset.} Each experiment is averaged over 6 trials.}
     \vspace{-1mm}
	\label{tab:coco-c}
    \begin{tabular}{l|c|c|c}
        \shline
        Method & mAP$_\texttt{clean}$ & mPC & rPC \\
        \shline
        Source (Retina-R50) & 37.4 & 18.3 & 48.9 \\
        FGD \cite{yang2021focal} &  39.6 & 20.3 & \underline{51.3} \\
        DiffKD \cite{huang2023knowledge} &  \underline{39.7} & \underline{20.3} & 51.1 \\
        \rowcolor{mygray}
	    FreeKD (Ours) & \textbf{39.9} & \textbf{20.8} & \textbf{52.1} \\
        \shline
	\end{tabular}
     \vspace{-1mm}
\end{table}
We evaluate the robustness of student detector RetinaNet-R50, trained with FreeKD on the COCO-C dataset \cite{michaelis2019benchmarking}. COCO-C is derived from \texttt{val2017} set of COCO, enriched with four types\footnote{including noise, blurring, weather, and digital corruption.} of image corruption, and each type further comprises several fine-grained corruptions. The results on corrupted images compared in Table \ref{tab:coco-c}, the mPC improvement of FreeKD compared to DiffKD \cite{huang2023knowledge} is greater than mAP$_\texttt{clean}$, and FreeKD outperforms DiffKD \cite{huang2023knowledge} by $1.0\%$ rPC\footnote{rPC = mPC / mAP$_\texttt{clean}$}. Our method is beneficial to enhancing the extra robustness and domain generalization abilities of the student.

\subsection{Large-Scale Vision Models Distillation}
To fully investigate the efficacy of FreeKD, we further conduct experiments on much stronger large-scale teachers.

\definecolor{mygray}{gray}{.9}
\begin{table}
  \centering
    \setlength{\tabcolsep}{2.5pt}
    \renewcommand{\arraystretch}{1.5}
    \footnotesize  
  \caption{\textbf{The performance of DETR-like models via FreeKD on COCO.} \textit{De-DETR}: Deformable DETR, \textit{MBv2}: MobileNetV2.}
  \begin{tabular}{c|cccccc}
    \shline
    Teacher & Student & Backbone & AP & AP$_{S}$ & AP$_{M}$ & AP$_{L}$ \\
    \shline 
      \multirow{4}{*}{\makecell{\scriptsize{De-DETR} \\ R101 \\ 47.1 (50e)}}  
      & De-DETR & \multirow{2}{*}{MBv2} & 33.5 & 16.9 & 36.4 & 46.6     \\
      &  + FreeKD &    & \cellcolor{mygray} \fg{36.2} (\fg{2.7$\uparrow$}) & \cellcolor{mygray} \textbf{19.3} & \cellcolor{mygray} \textbf{38.9} & \cellcolor{mygray} \textbf{49.0}\\
      \cline{2-7}
      & De-DETR & \multirow{2}{*}{R$18$} & 36.4 & 19.6 & 39.0 & 49.3   \\
      & + FreeKD &  & \cellcolor{mygray}\fg{38.9} (\fg{2.5$\uparrow$}) & \cellcolor{mygray} \textbf{22.0} & \cellcolor{mygray} \textbf{41.2} & \cellcolor{mygray} \textbf{51.9}\\
    \hline
    \multirow{4}{*}{\makecell{DINO \\ Swin-L \\ 56.6 (12e)}}
      & DINO & \multirow{2}{*}{R$50$} & 48.4 & 30.9 & 51.3 & 63.4     \\
      & + FreeKD &   & \cellcolor{mygray} \fg{50.4} (\fg{2.0$\uparrow$}) & \cellcolor{mygray} \textbf{33.1} & \cellcolor{mygray} \textbf{53.6} & \cellcolor{mygray} \textbf{64.9}     \\
      \cline{2-7}
      & DINO & \multirow{2}{*}{R$18$} & 45.1 & 28.7 & 48.0 & 59.1   \\
      & + FreeKD &   & \cellcolor{mygray} \fg{47.3} (\fg{2.2$\uparrow$}) & \cellcolor{mygray} \textbf{30.0} & \cellcolor{mygray} \textbf{50.4} & \cellcolor{mygray} \textbf{61.3} \\
    \shline
  \end{tabular}
  \label{tab:12ep}
\end{table}
\textbf{DETR-like Model.} For the object detection task, we apply FreeKD for two popular DETR-based models (Deformable DETR \cite{zhu2020deformable} and DINO \cite{zhang2022dino}) with various student backbones (R18, R50, and MobileNetV2 \cite{sandler2018mobilenetv2}). For De-DETR, FreeKD brings $2.5+$ AP improvement for both De-DETR-R18 and De-DETR-MBv2 students. While for DINO model, it still has a $2.0+$ AP gain for stronger students, e.g., DINO-R50 breaks the limit of $50$ AP with the help of FreeKD. Notably, we only distill the output of the final encoder layer and train the students in $12$ epochs (1X). 

\definecolor{mygray}{gray}{.9}
\begin{table}[t]
    \centering
    \setlength{\tabcolsep}{10pt}
    \renewcommand{\arraystretch}{1.5}
    \footnotesize
	\centering
	\caption{\textbf{The performance of SAM via FreeKD on SA-1B.}}
	\label{tab:seg_sam}
    \begin{tabular}{l|c|c|c}
        \shline
        Teacher & Students & Steps & mIoU \\
        \shline
        \multirow{3}*{\makecell{SAM \\ ViT-H}} & SAM ViT-Tiny & 20K & 40.12 \\
        ~ & + MSE & 20K & 42.42 \\
        ~ &  \cellcolor{mygray} + FreeKD & \cellcolor{mygray}20K & \cellcolor{mygray} \textbf{44.63} \\
        \shline
	\end{tabular}
     \vspace{-1mm}
\end{table}
\textbf{Segment Anything Model (SAM).} For the semantic segmentation task, SAM \cite{kirillov2023segment} is our first choice to validate the generality of FreeKD. We take the original SAM as the teacher, and its default image encoder is based on the heavyweight ViT-H \cite{dosovitskiy2020image}. Therefore, we replace the ViT-H with ViT-Tiny as the student and transfer the dark knowledge from image embeddings, which are generated by the image encoder. The student is trained with the SA-1B dataset \cite{kirillov2023segment} for 20K iterations (The image encoder is distilled for 10, 000 steps with 1024 × 1024 input size, and then the mask decoder is fine-tuned for 10, 000 steps) and evaluated with mIoU between the original SAM and itself. We run all the experiments on 8 A100 GPUs. For comparisons with our baseline one spatial-level feature distillation, we also report the mean square error (MSE) results with the same distillation location as FreeKD. As summarized in Table \ref{tab:seg_sam}, FreeKD obviously outperforms the MSE results by $2.21\%$ on SAM ViT-Tiny and improves the student by $4.51\%$. 

The above cases indicate that our precise frequency information in FreeKD is generic to large-scale vision models. Besides, sourced from Parameter-Efficient Fine-Tuning, the Prompt-guided distillation method thus is more fit for foundation vision teacher models, and effectively polishes up the performance of the students.
\section{Analysis}
\begin{figure}[t]
    \centering
    \includegraphics[width=0.48\textwidth, height=5cm]{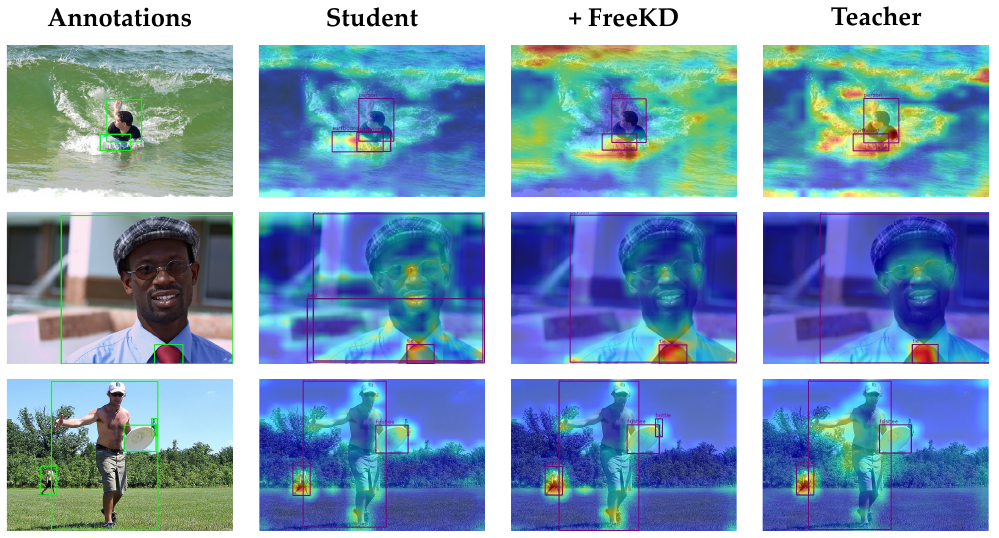}
    \caption{\textbf{Visualization of student features, student distilled with FreeKD features and teacher features on COCO dataset.}  The cases are randomly selected from val set and the heatmaps are generated with AblationCAM \cite{ramaswamy2020ablation}.}
    \label{fig:res_vis}
    \vspace{-2mm}
\end{figure}
\subsection{Effects of Frequency Prompts}
\begin{table}[t]
    \caption{\textbf{Ablation study on Frequency Prompts (FP).} We use RepPoints-R50 student and RepPoints-X101 teacher on COCO with various frequency bands.} 
    \label{tab:ab_1}
    \begin{subtable}{.5\linewidth}
	\renewcommand\arraystretch{1.2}
	\setlength\tabcolsep{1.mm}
	\centering
            \small 
        \begin{tabular}{cc|c}
        \shline
        \multicolumn{2}{c|}{Frequency Bands}  & AP \\
        \hline
        \multicolumn{3}{c}{\textit{Distill w/o FP.}}\\
        \hline
        Low & High &  \\
        \cmark & \xmark & 40.7 \\
        \xmark & \cmark & \textbf{41.8}   \\
        \cmark & \cmark & 41.3   \\
        \shline
	\end{tabular}
    \end{subtable}%
    \begin{subtable}{.5\linewidth}
        \renewcommand\arraystretch{1.2}
        \setlength\tabcolsep{1.mm}
         \centering
            \small 
        \begin{tabular}{cc|c}
        \shline
        \multicolumn{2}{c|}{Frequency Bands}  & AP \\
        \hline
        \multicolumn{3}{c}{\textit{Distill w/ FP.}}\\
        \hline
        Low & High &  \\
        \cmark & \xmark & 41.0 \\
        \xmark & \cmark & 42.3   \\
        \cmark & \cmark & \textbf{42.4}   \\
        \shline
	\end{tabular}
    \end{subtable} 
\end{table}
We propose a semantic Frequency Prompt (FP) to localize the PoIs of both high and low-frequency bands to compensate for their own limitations in the distillation. Here we conduct experiments to compare the effects of FP on different frequency bands in Table \ref{tab:ab_1}. We can see that: \textbf{(a)} Only low-frequency distillation cannot help polish up the student, and even impair the performance ($-0.5$ AP) when combined with high-frequency bands. \textbf{(b)} When Frequency Prompt provides accurate PoIs, the low-frequency band eliminates harmful samples with $0.3$ AP gain, and the high filters extra noise by $0.5$ AP improvement. \textbf{(c)} In general, FP has improved frequency distillation by $0.6$ AP and unified the distillation framework of frequency bands.

\subsection{Effects of Position-aware Weight}
\begin{table}[t]
    \renewcommand\arraystretch{1.8}
    \setlength\tabcolsep{1.5mm}
    \centering
    \caption{\textbf{The comparison of attention weights on COCO (AP) via FreeKD.} Teacher: RepPoints-X101. Student: RepPoints-R50.}
    \vspace{-1mm}	
    \label{tab:ab_2}
    \small
	\begin{tabular}{c|cccc}
	    \shline
            Student & SE  & Non-local & CBAM & Ours  \\
	    \shline
            37.4 & 42.2 & 41.9 & 42.1 & \textbf{42.4} \\
	    \shline
	\end{tabular}
\end{table}
To validate our Position-aware weight effectiveness, we choose several spatial attention (Squeeze and Excitation (SE) \cite{hu2018squeeze}, Non-local Module \cite{wang2018non}, and Convolutional Block Attention Module (CBAM) \cite{woo2018cbam}) to watch Frequency distillation. The results are reported on Table \ref{tab:ab_2}. We find that enhancing frequency distillation from channel dimension is a more effective method (SE and ours), compared with the other two. Besides, our position-aware weight includes distinguished object information with multi-scale receptive fields, which is more urgent to the frequency domain.

\subsection{Effects of Frequency Transformation Manner}
\begin{table}[t]
    \renewcommand\arraystretch{1.5}
    \setlength\tabcolsep{1.8mm}
    \centering
    \caption{\textbf{Various Frequency Transformation Manners for FreeKD.} We use
RepPoints-R50 student and RepPoints-X101 teacher on COCO.}
    \vspace{-1mm}	
    \label{tab:ab_3}
	\small
	\begin{tabular}{cc|c}
	    \shline
            Method & Mother Function & AP  \\
	    \shline
            DCT & Cosine &  41.9 \\
            DFT & Sine and Cosine & 42.0 \\
            DWT & Wavelet & \textbf{42.4} \\
	    \shline
	\end{tabular}
     \vspace{-1mm}
\end{table}

In terms of which frequency transformation is more suitable for distillation, we conduct detailed experiments on three methods (Discrete Cosine Transform (DCT), Discrete Fourier Transform (DFT), and Discrete Wavelet Transform (DWT)). As shown in Table \ref{tab:ab_3}, DWT based on Wavelet is significantly superior to DCT and DFT, whose mother functions are trigonometric functions. The reason is that wavelet provides frequency domain information at different scales, facilitating the analysis of local signal features, while the trigonometric function only provides global frequency domain information.

\subsection{Visualization}

We visualize the prediction results and heatmaps of the detector in Figure \ref{fig:res_vis} to further investigate the efficacy of FreeKD. We utilize RepPoints-R50 student and RepPoints-X101 teacher as an example. In general, FreeKD yields more clear contrast between low-frequency pixels and high-frequency pixels in heat maps, and it provides more distinctive observation. For instance, in the third case, the student trained by FreeKD performs better than the teacher (e.g., it detects the bottle successfully). The reason is that: Firstly, the high-frequency imitation principle in FreeKD, with shorter wavelengths, enables the student to effectively capture details, edges, and richer textures, thereby focusing on small but crucial regions. Secondly, position-aware relational loss enhances the student's sensitivity to positional information, particularly for small objects. Lastly, the combination of ground truth and soft labels provides the student with robust supervision signals.

\begin{figure}
    \centering
    \includegraphics[width=0.48\textwidth, height=2.6cm]{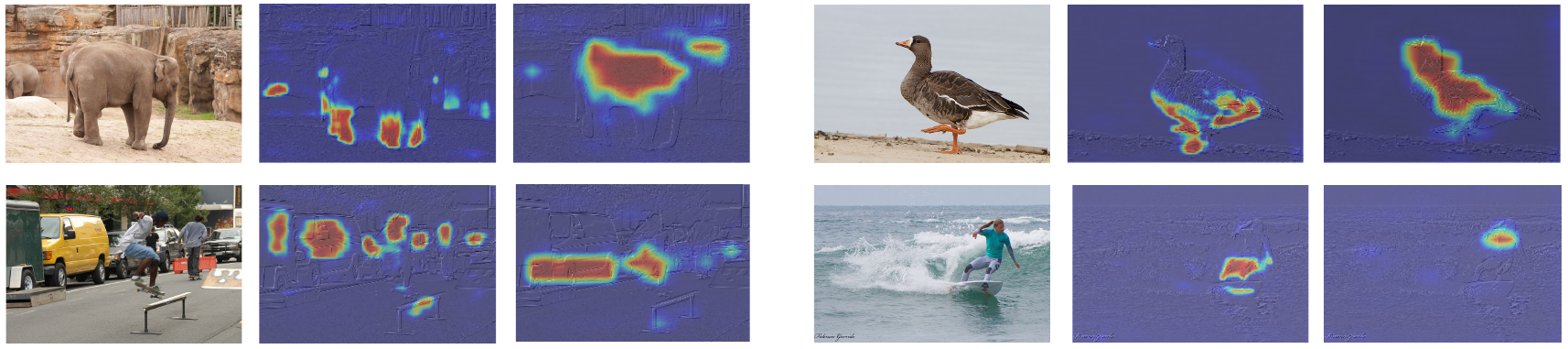}
    \caption{\textbf{Visualization of high-frequency pixels of interests on COCO dataset via RepPoints-X101.}}
    \label{fig:mask_vis}
    \vspace{-1mm}
\end{figure}

Meanwhile, FreeKD effectively avoid generating redundant bounding boxes in the first two cases, due to its spatial perception of objects.
Besides, we visualize the two PoIs (masks) generated by frequency prompt in the high-frequency band \texttt{HH} in Figure \ref{fig:mask_vis}. We find that the distinctive details in the band are marked out, while the noise is avoided to prevent performance degradation. This verifies our frequency prompt is effective in practice.
\section{Conclusion}
This research shifts the attention to frequency domain, and highlights its potential for knowledge distillation on dense prediction tasks. Meanwhile, to tackle the natural shortcomings of high and low frequency during mimicking, we introduce a novel pipeline named FreeKD, which determines both the optimal localization and extent for the frequency distillation. Specifically, we first propose Frequency Prompt to generate pixel-wise imitation principles. Besides, we design a channel-wise position-aware relational loss to enhance the sensitivity to objects for dense prediction. Extensive experiments demonstrated that FreeKD outperforms spatial-based distillation methods and provides more robustness to the student model.

\clearpage

{
    \small
    \bibliographystyle{ieeenat_fullname}
    \bibliography{main}
}

\clearpage
\setcounter{page}{1}
\maketitlesupplementary

\appendix

\section{Implementation Details}
\subsection{Object Detection}
We train the student with our FreeKD loss $\Ls_\mathrm{FreeKD}$, regression KD loss, and task loss for the object detection task. We set FreeKD loss weight to $1$ and regression loss weight to $1$ on \textit{Faster RCNN} students. For other detection frameworks, we simply adjust the loss weight of FreeKD to keep a similar amount of loss value as \textit{Faster RCNN}. Concretely, the loss weights $\mu$ of $\Ls_\mathrm{FreeKD}$ in Eq.\ref{eq:overall_loss} on \textit{RetinaNet}, \textit{FCOS}, and \textit{RepPoints} are $5$, $10$, and $10$.
\subsection{Semantic Segmentation}
For the segmentation task, we train the models with standard data augmentations including random flipping, random scaling in the range of $[0.5, 2]$, and a crop with size $512\times512$. The student is supervised by the FreeKD loss and task loss. Specifically, the loss weights $\mu$ of $\Ls_\mathrm{FreeKD}$ on \textit{PSPNet-R18} and \textit{DeepLabV3-R18} are $5$ and $5$, respectively.
\subsection{Distill on Segment Anything Model}

Segment Anything Model (SAM) \cite{kirillov2023segment} is a large-scale segmentation model characterized by the prompt proposed by Meta. To conduct distillation experiments on the SAM, we build a training framework based on the official code base\footnote{\url{https://github.com/facebookresearch/segment-anything}} and refer to the distillation pipeline in MobileSAM\footnote{\url{https://github.com/ChaoningZhang/MobileSAM}}. The pipeline is divided into two stages: first, distill the image encoder, and then fine-tune the mask decoder with the image encoder frozen. During FreeKD distillation, we utilize a full SA-1B dataset consisting of around $10$ million images to train the student model SAM-ViT-Tiny. The goal is to distill the student image encoder, with the officially released SAM-ViT-H model serving as the teacher. The image encoder is distilled for $10,000$ steps with $1024 \times 1024$ input size, and then the mask decoder is fine-tuned for $10,000$ steps. To speed up the training process, we simplify the finetune mask decoder process appropriately (e.g., only one round of interaction), and other settings strictly follow the original SAM for reproduction. We run all the experiments on 8 A100 GPUs.

\section{DWT meets Spatial-based Method}
\definecolor{mygray}{gray}{.9}
\begin{table}[t]
    \centering
    \setlength{\tabcolsep}{2.5pt}
    \renewcommand{\arraystretch}{1.5}
    \footnotesize
	\centering
	\caption{\textbf{The performance of spatial-based distillation methods via frequency domain on COCO val set.} Evaluate models with average precision (AP).}
	\label{tab:material}
    \begin{tabular}{p{1.2cm}p{1.3cm}|>{\centering\arraybackslash}m{2.3cm}|>{\centering\arraybackslash}m{2.3cm}}
        \shline
        \multicolumn{2}{c|}{Method} &  Spatial Domain & Frequency Domain \\
        \shline
        \multicolumn{2}{l|}{T: RepPoints-X101 } & \centering 44.2  & - \\
        \hline
        \multicolumn{2}{l|}{S: RepPoints-R50 }  & \centering 38.6  & - \\
        CWD \cite{shu2021channel} & \texttt{\scriptsize{ICCV21}} & \centering 41.8 &  42.0 \\
        PKD \cite{cao2022pkd} & \texttt{\scriptsize{NeurIPS22}} & \centering 42.0 &  42.1 \\
        \rowcolor{mygray}
	    FreeKD & & \centering - & \textbf{42.4}   \\
        \shline
	\end{tabular}
     \vspace{-2mm}
\end{table}
To investigate the feasibility of directly applying other spatial-based distillation methods to frequency domain distillation, we conduct experiments using CWD \cite{shu2021channel} and PKD \cite{cao2022pkd} as examples. For CWD, we minimize the Kullback–Leibler (KL) divergence between the channel-wise probability map of the high-frequency bands in the teacher and student. To apply PKD to the frequency domain, we try to reduce the representation gap between teacher and student via normalizing the high-frequency features with Pearson correlation. The results are listed in Table \ref{tab:material}. We find that transferring the distillation method from the spatial domain to the frequency domain further improves the accuracy of the student model. Meanwhile, we notice that the convergence speed and training speed of frequency domain distillation are both higher than spatial domain distillation. This is because neural networks initially learn low-frequency information and later focus on high-frequency context, which allows them to mimic the high-frequency components of the teacher model as a form of previewing.

\section{Confusion Matrix with FreeKD on COCO}
\begin{figure}[t]
    \centering
    \includegraphics[width=0.48\textwidth]{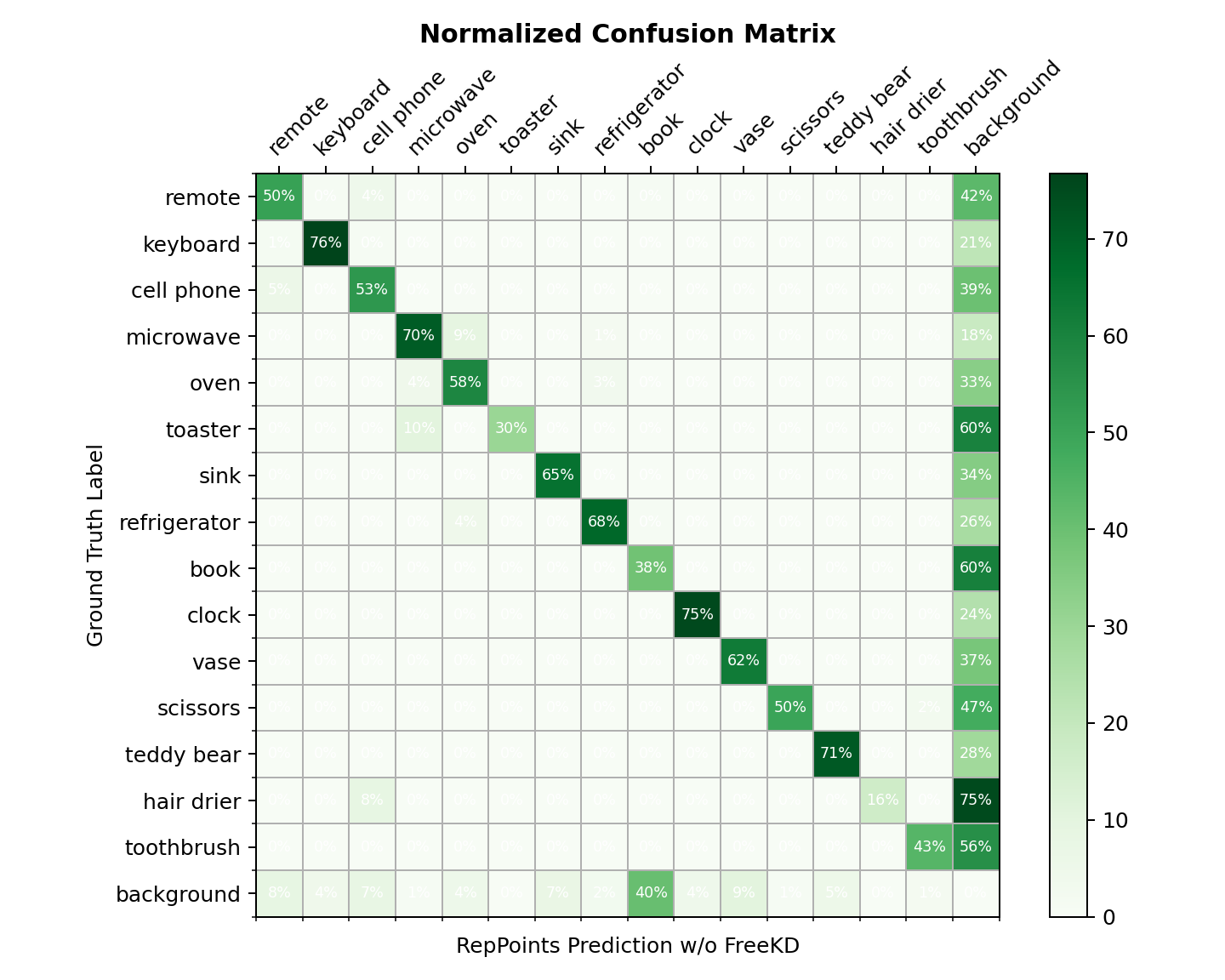}
    \includegraphics[width=0.48\textwidth]{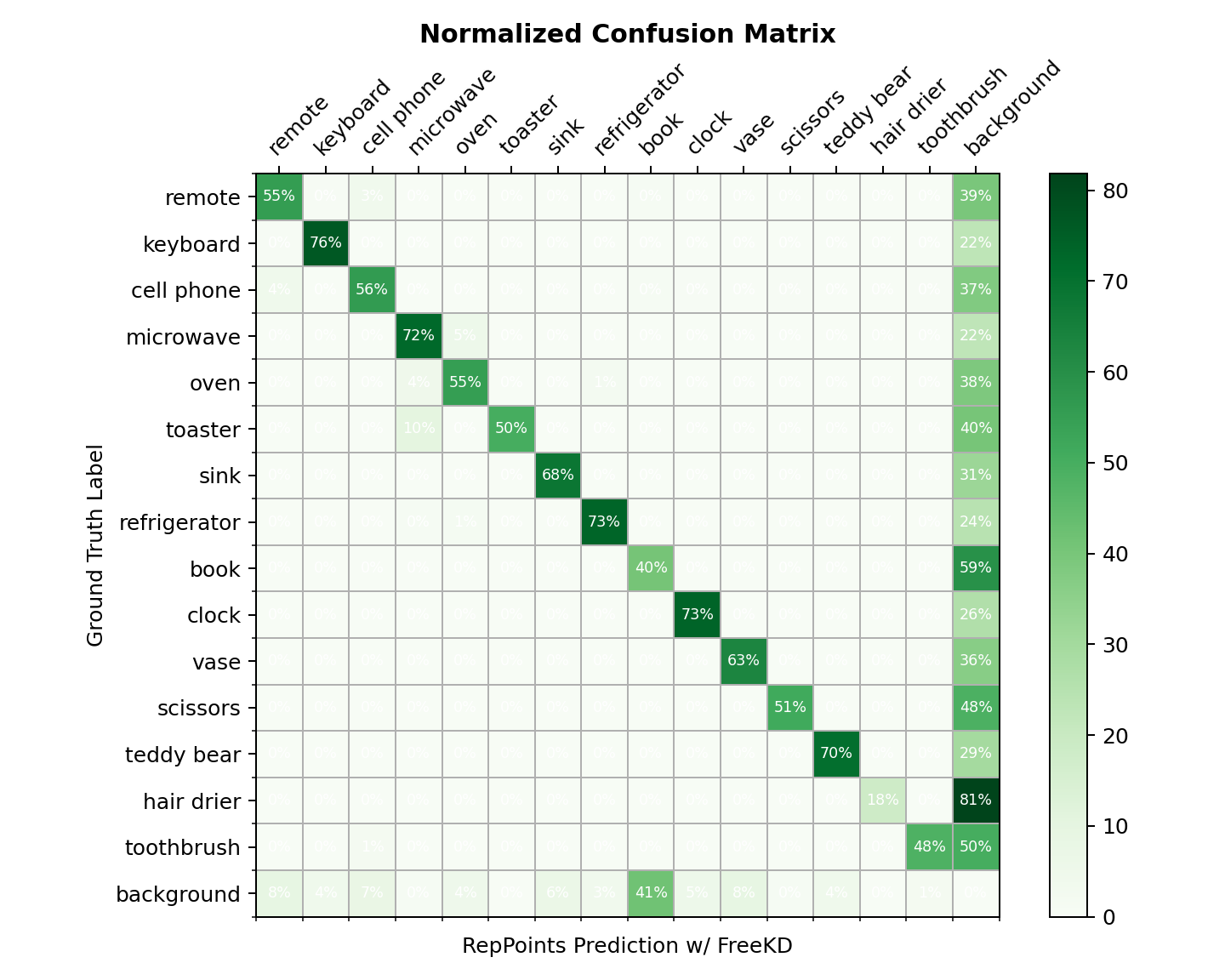}
    \caption{\textbf{Confusion matrix between the original student predictions and the student distilled by FreeKD predictions.}}
    \label{fig:res_con}
\end{figure}

We compute the confusion matrix of our method and make a comparison with the vanilla student in Figure \ref{fig:res_con}. We utilize RepPoints-R50 student and RepPoints-X101 teacher as an example. The normalized values on the diagonal of the confusion matrix represent the ratio of predictions that match the ground truth predictions. Our method achieves a higher ratio of matching in most cases (e.g. $20\%$ improvement on the toaster), which further validates that our method could transfer more knowledge from the frequency domain.

\section{More Visualizations}

\subsection{Visualization of Feature Maps}

\begin{figure}[t]
    \centering
    \includegraphics[width=0.48\textwidth]{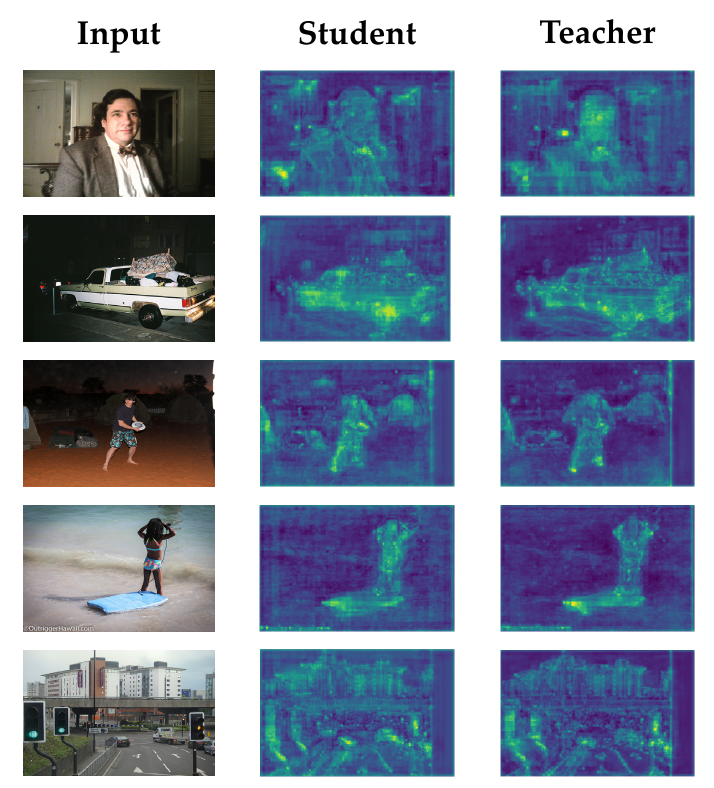}
    \caption{\textbf{Visualizations of the distilled student features and teacher features on
COCO dataset.} We utilize RepPoints-R50 student and RepPoints-X101 teacher as an example.}
    \label{fig:vis_fea}
\end{figure}

We visualize the features of student and teacher models in the first output (stride 4) of FPN in Figure \ref{fig:vis_fea}. The models used to extract the feature are RepPoints-R50 (student) and RepPoints-X101 (teacher) trained on COCO dataset. Following DiffKD~\cite{huang2023knowledge}, we average the feature map along the channel axis and perform softmax on the spatial axis to measure the saliency of each pixel. Formally, with a given feature map $\bm{X}\in\mathbb{R}^{C\times HW}$, we first average the channels and get $\bm{X}'\in\mathbb{R}^{HW}$, where
\begin{equation}
    X'_i = \frac{1}{HW}\sum_{i=1}^{HW}(\bm{X}_{:,i}),
\end{equation}
Then we generate the attention map for visualization as
\begin{equation}
    \bm{V} = H \cdot W \cdot softmax(\bm{X}' / \tau),
\end{equation}
where $\tau$ is the temperature factor for controlling the softness of distribution, and we set $\tau=0.5$.

\subsection{Visualization of Frequency PoIs}
\begin{figure}[t]
    \centering
    \includegraphics[width=0.48\textwidth]{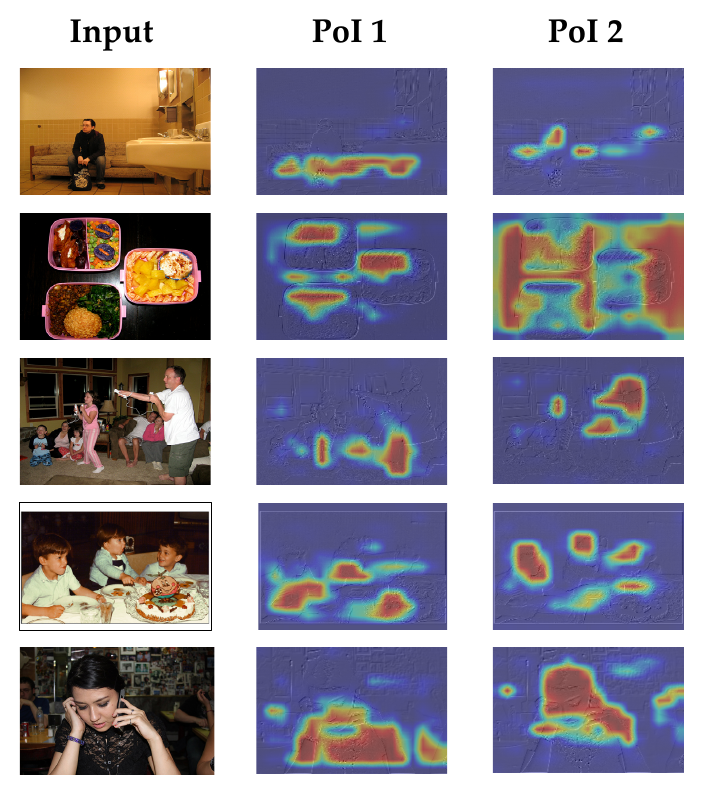}
    \caption{\textbf{Visualizations of Frequency PoIs in high-frequency band \texttt{HH} on
COCO dataset.} We employ the RepPoints-X101 teacher to generate the Frequency Prompt.}
    \label{fig:vis_FRE}
\end{figure}

We visualize the more two PoIs (masks) generated by frequency prompt in the high-frequency band HH in Figure \ref{fig:vis_FRE}.

\end{document}